# Academic Journals' AI Policies Fail to Curb the Surge in AI-assisted Academic Writing

Yongyuan He[1] and Yi Bu[1,2,*]

[1] Department of Information Management, Peking University, Beijing 100871, China

[2] Center for Informationalization and Information Management Research, Peking University, Beijing 100871, China

*Corresponding author. Email: buyi@pku.edu.cn

**The rapid integration of generative AI into academic writing has prompted widespread policy responses from journals and publishers. However, the effectiveness of these policies remains unclear. Here, we analyze 5,114 journals and over 5.2 million papers to evaluate the real-world impact of AI usage guidelines. We show that despite 70% of journals adopting AI policies (primarily requiring disclosure), researchers' use of AI writing tools has increased dramatically across disciplines, with no significant difference between journals with or without policies. Non-English-speaking countries, physical sciences, and high-OA journals exhibit the highest growth rates. Crucially, full-text analysis on 164k scientific publications reveals a striking transparency gap: Of the 75k papers published since 2023, only 76 (~0.1%) explicitly disclosed AI use. Our findings suggest that current policies have largely failed to promote transparency or restrain AI adoption. We urge a re-evaluation of ethical frameworks to foster responsible AI integration in science.**

## Introduction

In recent years, generative AI like ChatGPT has rapidly reshaped the academic research environment (1–10). These tools are increasingly used in academic writing, e.g., translation, summarization, and language polishing (11–13). While enhancing research efficiency (14) and breaking down language barriers (15) to promote global knowledge sharing (16, 17), AI also raises concerns about hallucinations (18), potential ethical issues, and controversies over whether AI should be credited as an author (19). To address these challenges, in 2023, authoritative organizations like the International Committee of Medical Journal Editors (20) and Committee on Publication Ethics (21) established guidelines that exclude AI from authorship while requiring researchers to clearly declare AI usage. Major publishing groups including Springer Nature, Elsevier, and Wiley have updated their submission guidelines to specify acceptable AI use, disclosure requirements, and attribution of responsibility. For example, *Science* allows authors to use AI for writing assistance but requires clear declaration in submissions (22).

However, given the rapid development and widespread application of AI, the actual effectiveness of these policies remains unclear. According to *Nature*, about 14% of the 1.5 million biomedical research abstracts indexed in PubMed in 2024 contained signed words of large language model (LLM)-generated text (23), showing that AI's influence in academic writing has reached a considerable scale. Although relevant policies have been implemented for some time, current research still lacks sufficient analysis of policy effectiveness. Facing this impact brought by AI, some questions urgently need exploration: Do these policy measures aimed at regulating AI use truly play their expected regulatory role? To what extent do they influence researchers' actual writing behavior?

To answer these questions, we conducted analyses of AI usage policies across 5,114 journals from the Journal Citation Report Q1 category and their 5,235,012 papers published between January 2021 and June 2025. We leveraged LLMs to classify journal guidelines and employed maximum likelihood estimations (MLE) combined with multiple robustness checks to quantify the actual prevalence of AI-generated content. Additionally, we counted the number of papers among 164,579 full-text publications that disclosed AI use to examine compliance with AI disclosure policies. This analysis provides evidence for assessing the real effectiveness of current academic AI governance.

## Results

**The widespread adoption of AI usage policies.** Analysis of the 5,114 journals reveals a widespread implementation of AI governance. Based on the policies collected in January 2025, approximately 70% of the journals have established AI usage guidelines. We classified these policies into four categories (Fig. S1), namely strict prohibition (explicitly banning author AI use), open policy (allowing AI use without mandatory disclosure), disclosure required (permitting AI use but mandating declaration), and not mentioned (no identifiable policy). Representative policy examples for each category are provided in Table S1. Among the four policy categories, 3,556 journals require disclosure, 1,529 journals do not mention policies, 27 journals strictly prohibit, and 2 journals have open policies. Due to the small number of journals with open or strict prohibition policies, we simplified the categories into two groups, journals with and journals without AI policies. Journals with AI policies consistently outnumber those without them across academic disciplines, open access (OA) status, and major publishers (Figs. 1a-1c). Through both human review and LLM analysis, we further analyzed journals that require authors to report their AI usage. We found that most journals request authors to report their AI usage in sections such as methods and acknowledgments, while a few journals require reporting through the submission system or cover letters (Fig. 2a). Moreover, 96.8% of journals allow authors to use AI for writing and editing, and 62.9% permit language and grammar checking according to their stated policies (Fig. 2b).

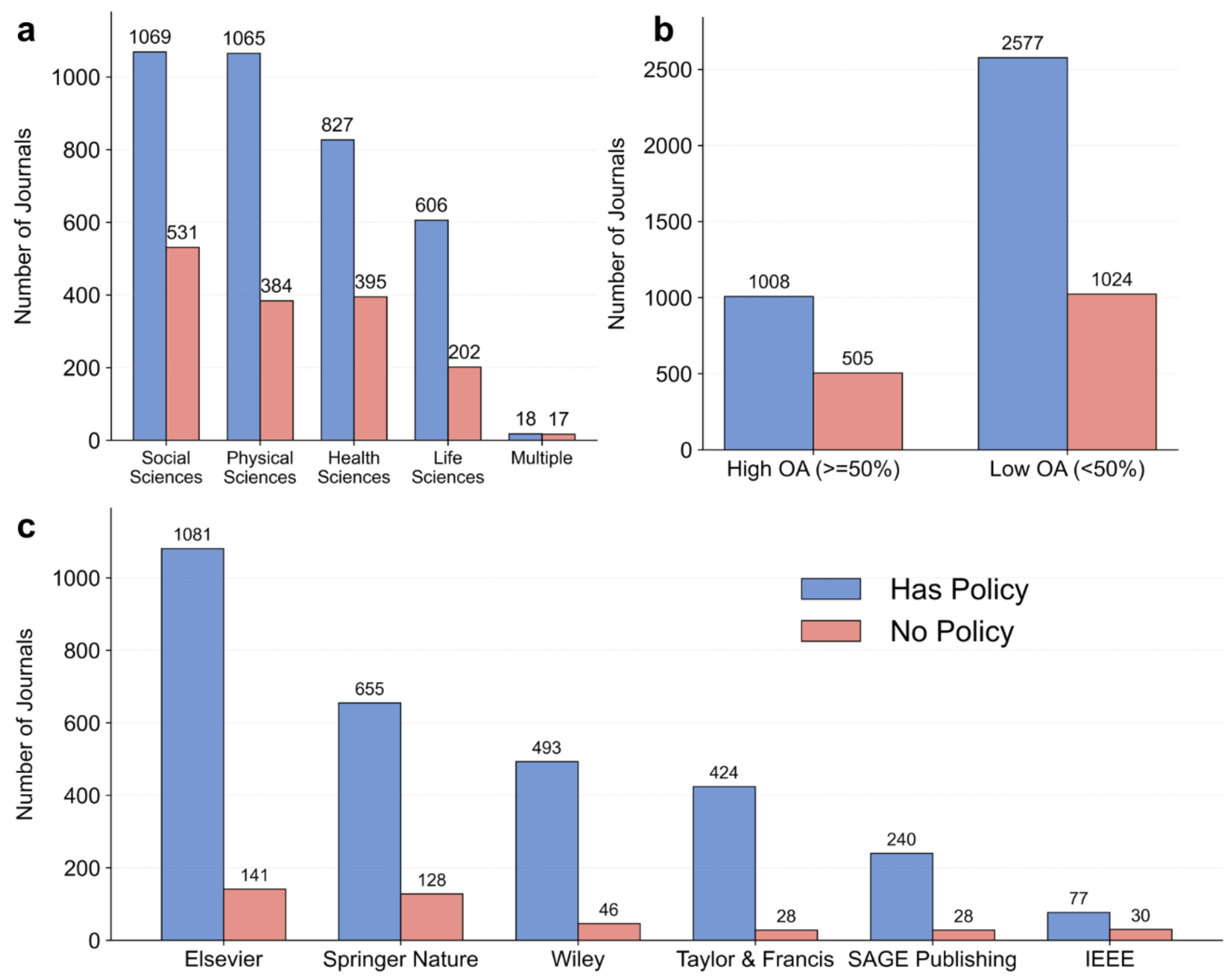

**Figure 1. Distribution of journals with/without AI policies across different categories. a**, Distribution by academic domains showing the number of journals with (blue) and without AI policies (red). Here, "multiple" refers to journals covering multiple domains. **b**, Distribution by open access (OA) status, comparing journals with high (≥50%) and low OA rates (<50%) (Materials and Methods). c, Distribution by major academic publishers.

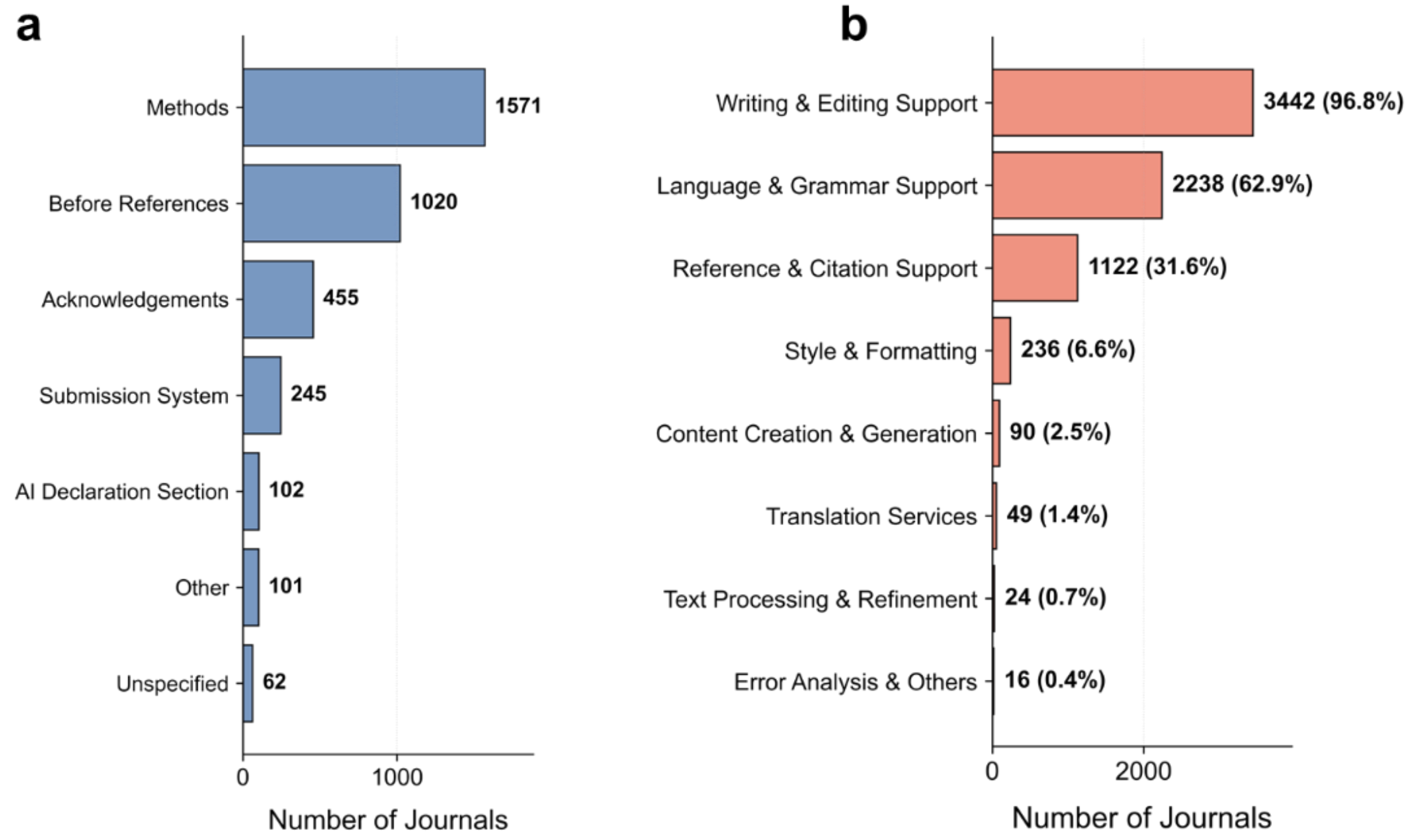

**Figure 2. Distribution of AI disclosure requirements and permitted AI usage categories among journals requiring AI disclosure. a**, Disclosure location requirements showing where journals mandate authors to report AI usage, with methods section being the most common location, followed by the section before references, acknowledgements, and other sections. **b**, Permitted AI usage categories among journals allowing AI use, with Writing & Editing Support being the most prevalent category, followed by Language & Grammar Support, Reference & Citation Support, and various other specialized applications.

In an updated set of policies collected in October 2025, we observed that the number of journals with disclosure required policies surged by nearly 800 (primarily converting from the "Not mentioned" category), whereas the number of "Strict prohibition" and "Open policy" policies remained very limited (Fig. S4). This trend indicates that AI governance in academic publishing is continuing to expand, maintaining a stance that accepts the use of AI tools while requiring disclosure.

**The limited impact of AI policies on academic writing.** Despite the high prevalence of AI policies, our results show a significant increase in AI content proportion from 2023 onward, indicating that academic writing has been greatly influenced by ChatGPT's release in November 2022. Notably, journals with and without AI policies show parallel growth trends (Fig. 3a), suggesting that publishing AI policy has not affected researchers' use of AI writing tools. Statistical analysis of the logarithmic monthly growth rates confirmed no significant difference between the two groups (Table S4; Materials and Methods). For robustness tests,  we used three additional detection methods to validate the reliability of our primary detection method: a dictionary-based AI keyword frequency analysis, the excess word analysis proposed by Kobak *et al.* (24), and an independent verification using the online detection tool ZeroGPT (25). Furthermore, a supplementary analysis of 164,579 full-text publications confirmed that these trends still hold regardless of using only abstracts or full texts of scientific publications (Materials and Methods; Fig. S3e).

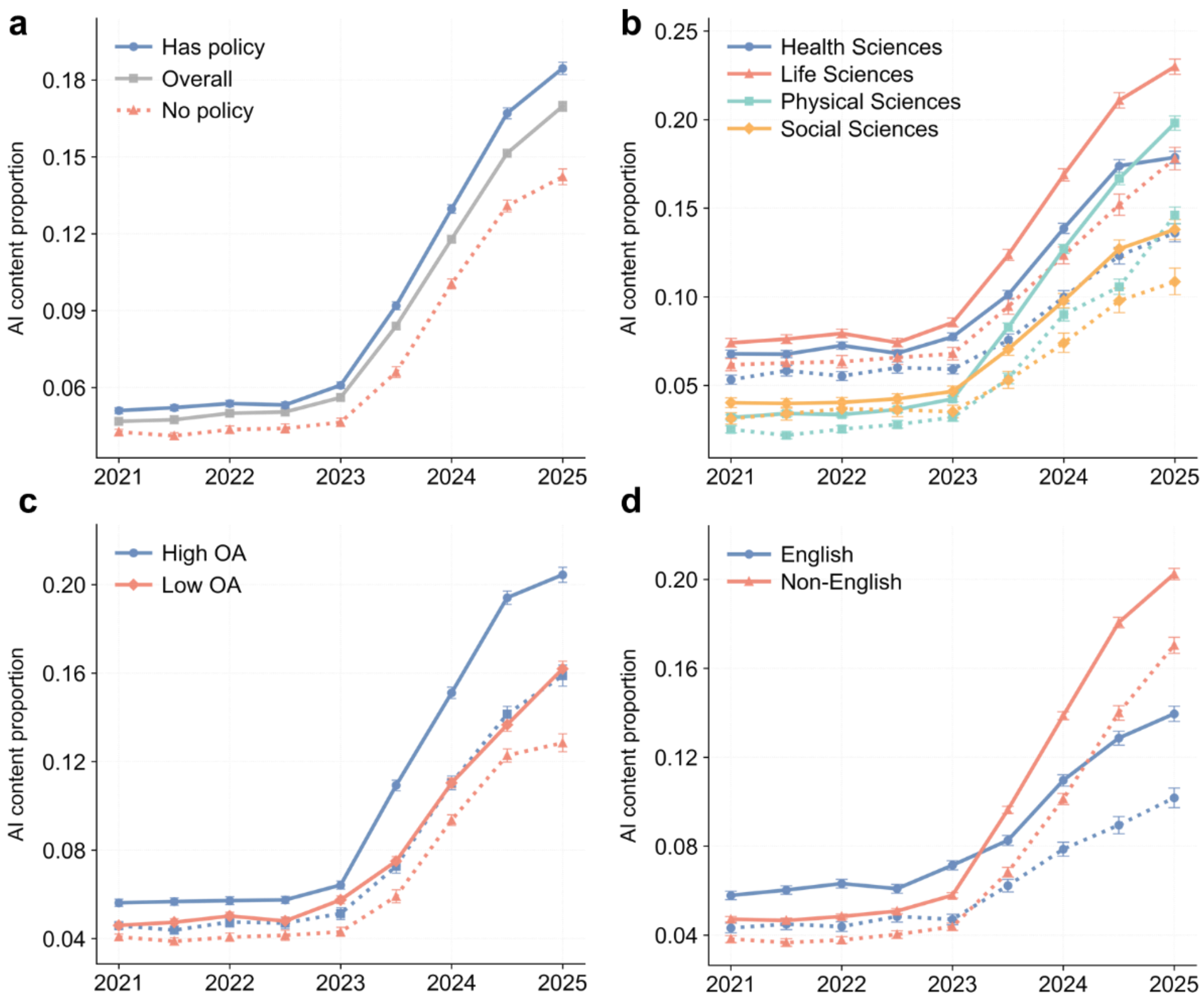

**Figure 3. Temporal trends in AI content proportion across different journal categories (2021–2025). a**, Trends by AI policy status showing journals with/without AI policies and the overall average. **b**, Trends by academic domains, with solid lines representing journals with AI policies and dashed lines representing journals without AI policies. **c**, Trends by OA status. **d**, Trends by authors' country classification comparing English- and non-English-speaking countries (Materials and Methods). Error bars represent 95% confidence intervals.

**Disparities in AI tool adoption across domains, countries and open access status.** At the domain level, physical sciences showed the most significant increase in AI content proportion, suggesting this domain has been potentially most influenced by AI writing tools (Figs. 3b and S3a). In contrast, social sciences showed relatively smaller increases, possibly due to the requirements in writing methods or research ethics, resulting in lower dependence on or acceptance of AI. At the country level, we divided all countries into two groups according to their authors' affiliation information: English- (e.g., U.S., U.K., Australia, and Canada) and non-English-speaking (e.g., China, Germany, Japan, and Brazil) countries. Data shows that after 2023, non-English-speaking countries experienced faster growth in AI content proportion than English-speaking countries (Fig. 3d), with China showing the most prominent increase (Fig. S3b). We then classified all papers into low and high OA groups based on the journals in which they were published and found that the high OA group experienced faster growth (Materials and Methods; Fig. 3c). This trend is corroborated at the publisher level: fully open-access publishers (e.g., MDPI and Frontiers) exhibited higher AI content proportion compared to publishers with lower OA rates, such as Elsevier, Springer Nature, and Wiley (Fig. S5). Across all domains, countries, and OA status, the growth patterns between journals with and without AI policies remained similar, reinforcing our finding that policy presence does not change AI adoption trends. Specifically, the logarithmic monthly increments of the AI content proportion did not differ significantly between the groups with and without AI policies across three categories (Table S4).

**Disclosure of AI usage in full-text analysis.** Our analysis based upon 164,579 full-text scientific publications shows that, among the 75,172 papers published since 2023, only 76 papers (~0.1%) disclosed using AI tools for writing assistance in the methods and/or acknowledgment sections. However, the disclosure rate has shown a consistent upward trend, rising from 0.01% in early 2023 to 0.43% in the first quarter of 2025 (Fig. S6). Despite this growth, transparency lags dramatically behind actual adoption. By calculating the ratio of AI content proportion to disclosure rate, we find an underreporting phenomenon. For instance, in the first quarter of 2025, the ratio was approximately 40:1. In other words, for every 40 papers showing statistical evidence of AI usage, only one formally disclosed it. Additionally, 2/3 of these disclosed papers were published in journals with AI policies (Fig. 4b), and most disclosed their AI usage in acknowledgments (Fig. 4c). Furthermore, ChatGPT was the most frequently disclosed tool (Fig. 4d), with AI primarily employed for writing and editing tasks, such as general writing assistance, general editing, and readability improvement (Fig. 4e).

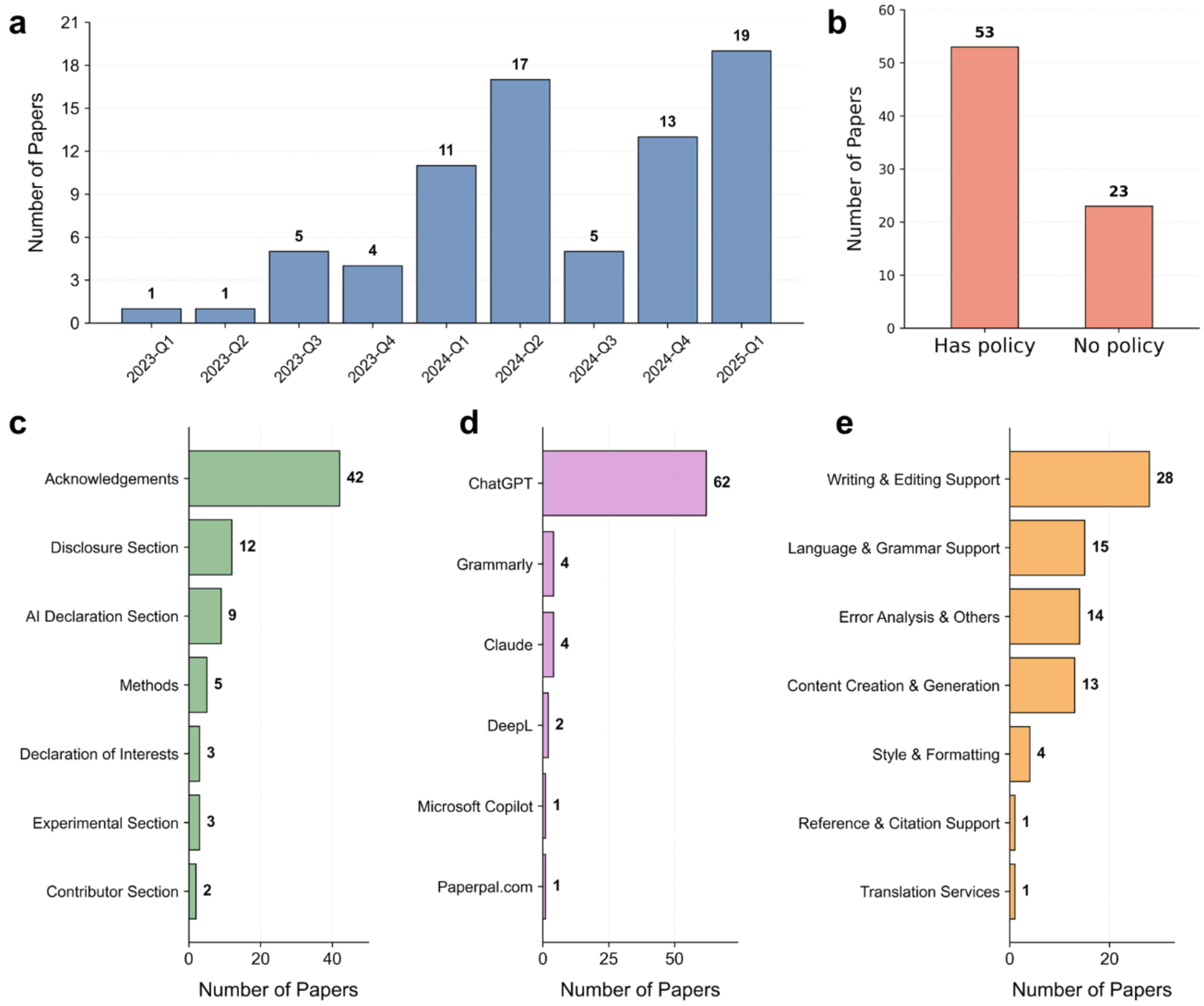

**Figure 4. Analysis of AI usage disclosure in scientific publications through full-text analyses. a**, Temporal distribution of papers with detected AI usage disclosure from 2023-Q1 to 2025-Q1. **b**, Distribution of papers with detected AI disclosure by journal AI policy status. **c**, Distribution of AI usage disclosure locations within papers. **d**, Distribution of specific AI tools mentioned in disclosure statements. **e**, Distribution of disclosed AI usage purposes categorized into functional types.

## Discussion

The release of ChatGPT in November 2022 has been reshaping the way of science communication, leading to a widespread and sustained increase in AI-assisted writing. Although this growth trend is universal, the rate of increase varies significantly across different groups. Specifically, that growth has been particularly rapid in the physical sciences, among papers from high OA journals and among researchers from non-English speaking countries. For instance, extant studies have shown that authors from non-English speaking countries are more likely to rely on AI tools compared with their English speaking countries' counterparts (17, 24, 26–28), reinforcing the view that AI plays an important role in dismantling language barriers in scholarly communication (29–31). Regarding disciplinary differences, prior survey have documented marked differences in familiarity and usage across broad disciplinary areas (32). Specifically, applied and technology-oriented disciplines report higher levels of AI knowledge and use than arts, humanities, and many social science disciplines (33, 34). Our empirical findings supplement existing works by illustrating that journals with greater OA rates exhibit a higher level of AI content proportion, which is consistent with Kobak *et al.* (12). The publishing model characterized by rapid review, streamlined processes, and being APC-driven might be accompanied with an increasing risk of publishing papers with greater proportion of AI content (35).

Notably, our results show no significant difference in the growth rate of AI usage between journals with AI policies and those without, indicating that the current policy has not yet effectively slowed the adoption of AI technology. However, the inability of policies to curb the proliferation of AI does not mean they are without value. Our full-text analysis reveals that, although the base number remains low, the number of papers explicitly disclosing AI use has shown a gradual increasing trend since 2023. This suggests that policies may be guiding a subset of researchers from use toward declaration, gradually fostering more transparent academic practices. Therefore, the true objective of these policies may not be to prevent the use of AI, which is neither practical nor necessary, but to regulate its application and prevent potential academic misconduct (36). By clarifying that human authors bear full responsibility for AI-generated content and prohibiting extreme cases of entirely AI-generated text, these policies help shape a new research paradigm of human-machine collaboration, one that fully leverages AI assistance while ensuring that human authors remain central to the process of knowledge creation.

Most journals do not explicitly prohibit the use of AI tools, but, instead, focus on requiring transparent disclosure and specifying that AI cannot be listed as a co-author. That being said, what remains a direction for future research is whether authors actually disclose their use of AI tools. In our full-text sample, the fact that 0.1% of the publications since 2023 explicitly acknowledged their AI use may stem from multiple factors. First, authors' awareness and understanding of these newly implemented policies remain limited and uneven, and there is still ambiguity regarding the extent to which usage requires formal disclosure. Second, authors may be concerned about how such disclosures will be perceived by editors, reviewers, and the wider scientific community. They might worry that admitting to the use of AI could cast doubt on the originality of their intellectual contributions, potentially leading to stricter scrutiny, bias during peer review, or negative impact on their reputation (37–40). Furthermore, academia has yet to establish a consensus on norms governing the use of AI, and relevant policies are still evolving, which contributes to authors' cautious approach in deciding whether and how to disclose their use of AI.

Facing the profound impact of AI, simple prohibition or disclosure policies are insufficient to address this technological wave. Future challenges include developing smarter tools to verify the originality and accuracy of academic content and encouraging transparent and ethical use of AI (41). We have seen positive practical attempts, such as Stanford's "Open Conference of AI Agents for Science" (42) which innovatively requires AI to be listed as first authors. Simultaneously, we must recognize AI's transformative potential in promoting academic fairness and knowledge dissemination. What we need is not opposition and resistance, but proactive engagement and institutional innovation to ensure AI technology truly enhances the value of science.

We acknowledge several limitations in this pilot study. First, regarding the accuracy of AI detection, the MLE method provides probabilistic estimates at the group level rather than definitive judgments on individual manuscripts. Although we validated our results using multiple approaches, these operationalizations may still yield false positives, particularly when distinguishing between AI-generated text and the formulaic writing styles in certain technical disciplines or the linguistic patterns typical of non-native English speakers. Second, while our detection method can estimate the AI content proportion, it would be quite challenging to precisely discern the specific extent or nature of AI involvement. Specifically, we are unable to distinguish whether AI is used merely for linguistic polishing which is typically permissible under current policies, or for substantive text generation that may violate regulations. This technical limitation prevents us from verifying the effectiveness of specific policy details (e.g., determining whether "polishing-only" policies effectively curb "ghostwriting"). Consequently, our policy analysis is confined to a binary classification of "with/without policy", precluding a deep evaluation of specific policy content. Future research needs to develop more refined detection frameworks or integrate qualitative surveys to distinguish compliant assistive editing from non-compliant content generation. Third, concerning data scope, while our abstract-based analysis covers over 5.2 million papers, our full-text analysis was limited to a sample of 164k publications due to copyright and technical constraints.

We acknowledge a potential selection bias in this subset: papers from high-OA publishers were more accessible for retrieval. Finally, in terms of observation windows, we should note that the policy data were collected in two batches in January and October 2025, respectively, and the current release dates of the policies for each journal are lacking. Therefore, some policies may have been released very recently and have not had sufficient time to exert a substantive impact in academic publishing practices. Considering that the academic publishing process involves several months or even longer, there is a certain time lag from the introduction of a policy to the manifestation of its effects in published papers. Consequently, our results do not necessarily imply that the policies are ineffective, but their effects have not yet been fully reflected in the data. Assessing their long-term effects requires tracking over a longer period.

## Materials and Methods

**Data collection and processing.** Data acquisition involved two primary streams. First, journals' AI policies were aggregated through a hybrid approach: a manual review in January 2025 followed by an automated web-crawling pipeline in October 2025, with LLMs employed for classification verification. Second, publication metadata were retrieved from OpenAlex (43) (covering publications between January 2021 and June 2025). To address missing abstracts in the original dataset, we supplemented records using the Web of Science database. After merging and filtering for completeness, the final dataset comprised 5,235,012 publications. Detailed retrieval protocols and validation steps are provided in the Supporting Information.

**Policy classification.** We used GPT-4o-mini to identify and classify journal policies into four categories based on a well-designed prompt (see Supporting Information). Crucially, the classification focused strictly on generative text guidelines for authors; policies addressing only reviewer conduct or image manipulation were explicitly filtered out and categorized as "Not mentioned." To ensure robustness, we employed Gemini 2.5 Flash as an independent validator. Any discrepancies between the two models were subjected to manual adjudication. These conflicts primarily arose from nuanced policy language: Specifically, instances where a policy explicitly described itself as "open" or "liberal" but subsequently included a specific clause mandating disclosure. In such cases, we prioritized the operational requirement (the mandate to disclose), correcting the classification to "Disclosure required."

**Determination of country affiliation.** Based on the paper-author-institution-country mapping relationships recorded in the OpenAlex database, we adopted the following method to determine the country affiliation of publications. For each publication, we extracted the country information corresponding to all affiliated institutions of every author and removed duplicates, thereby establishing a direct mapping between publications and countries. In our statistical analysis, we employed a full counting method (44), meaning that each publication contributes one count to every country with which it is associated.

**Open access (OA) classification.** Journal Citation Report (JCR) does not provide explicit OA status indicators for journals. However, it includes a "% of Citable OA" field, which represents the percentage of OA articles among all citable publications within each journal. To classify journals into meaningful OA categories for our analysis, we established an approach: journals with ≥50% citable OA articles were classified as the High OA group, while those with <50% were classified as the Low OA group. We further obtained the OA status (recorded as either "True" or "False") of these journals from the OpenAlex database. As shown in Fig. S2, journals with a "% of Citable OA" greater than 50% have a higher probability of being classified as OA, whereas those below this threshold show a greater likelihood of being non-OA. This trend aligns with our pre-defined "High OA" and "Low OA" classification criteria based on JCR data. It also indicates a strong association between the proportion of OA articles in a journal and its overall OA status, showing the robustness of the empirical results.

**Academic domain classification.** The four academic domains referenced throughout this study (Health Sciences, Life Sciences, Physical Sciences, and Social Sciences) are based on the OpenAlex classification system (45). These domains represent broad academic categories, each encompassing a wide range of specific disciplines. For instance, the "Physical Sciences" domain is not limited to core subjects like physics, but also includes mathematics, computer science, and related engineering fields. However, JCR employs a more granular classification scheme with over 200 subject categories, which makes it quite challenging for domain analysis. To address this discrepancy, we mapped each JCR subject category to the most appropriate OpenAlex domain based on subject matter overlap, methodological similarities, and disciplinary traditions with the help of ChatGPT. This mapping process was conducted iteratively with human verification at each step to ensure accuracy.

**Statistical analysis.** To compare trends between groups (e.g., has policy vs. no policy), we calculated the logarithmic monthly increment of AI content proportion and assessed statistical significance using the Mann-Whitney U tests.

**AI detection methods.**

*Maximum Likelihood Estimation method.* Our primary method for detecting AI-generated content employs the MLE approach proposed by Liang *et al*. (46). This method estimates the probability that a given text contains AI-generated content by analyzing word frequency distributions. We define the core metric derived from this method as AI content proportion. Crucially, this metric is obtained through distributional analysis at the group level, representing the estimated probability of AI content presence across all samples within a specific category, rather than the directly measured percentage of AI-generated text in individual papers. The core assumption is that observed textual distributions can be represented as weighted mixtures of human-written and AI-generated text patterns. For each group's abstract, we extracted all adjectives, calculated their frequency distributions, and estimated the mixing parameter by maximizing the log-likelihood function to determine the AI content proportion.

*AI keyword frequency analysis.* Recent studies have shown that AI-generated academic texts often exhibit distinctive lexical patterns, which can be captured using lists of AI-characteristic words derived from previous corpora and AI-detection tools (47–49). Following these studies, we employed a keyword frequency approach to measure the extent of AI usage in text. First, we collected AI-characteristic vocabularies from four previous studies and two AI detection tool websites (Table S2), combined these sources, and selected 110 words with frequencies greater than two. This process was necessary because previous studies tended to focus on specific disciplines. To ensure accuracy across diverse academic fields, we collected vocabularies from different disciplines and sources, creating a more representative AI-characteristic word set. We applied a differentiated measurement strategy: abstracts were flagged as AI-modified if they contained at least one target keyword, whereas full-text analysis quantified the total frequency of these keywords per article. This keyword-based method yielded results consistent with the MLE approach (Fig. S2d) and showed significant positive correlation with MLE detection results ($r$=0.938, $p$<0.001) (Fig. S2f).

*Excess word analysis.* We employed the excess word analysis framework proposed by Kobak *et al*. (24) as a robustness check. This framework does not rely on any pre-defined AI lexicon, but rather identifies LLM-assisted writing by detecting style words that exhibit abnormal increases in frequency relative to a pre-LLM baseline period. Specifically, we used the lexical frequencies from 2021–2022 (not affected by LLMs) as a baseline to predict the expected word frequencies for subsequent years via linear extrapolation. The difference between the actual and expected frequencies serves as a lower-bound estimate of AI usage probability. Unlike the original study that used annual data to construct the baseline, our dataset covers a relatively short time window (2021–2025). To more accurately capture vocabulary usage patterns, we adopted a monthly granularity strategy: that is, we used data from the corresponding month in both 2021 and 2022 to predict the expected word frequency for the same calendar month in 2023–2025 (50). This refined monthly baseline model effectively mitigates potential estimation bias due to the limited temporal span of the data. The results show that the monthly AI usage rates derived from the excess word method for 2023–2025 are highly consistent with our primary MLE results ($r$=0.979, $p$<0.001), further confirming the robustness of our findings.

*Online AI detection tool.* Since aforementioned methods rely on word frequency statistics, and different academic fields may have distinct vocabulary patterns, we further validated our findings using the online AI detection tool ZeroGPT (25) as an independent verification method. Due to resource constraints, we sampled 30 abstracts from each month between 2021 and June 2025. The online detection results also showed positive correlation with our MLE results ($r$=0.657, $p$<0.001) (Fig. S2f), demonstrating the robustness of our findings. The low correlation might be attributed to the inherent design of online detection tools like ZeroGPT, which tend to flag only texts exhibiting high-confidence AI signatures (17) (typically observed in 2025 data). Consequently, these tools might lack sensitivity to the subtle or partial AI usage prevalent during 2023–2024; for lightly edited or hybrid texts, the output probability scores tend to plateau at lower levels, failing to capture the nuanced continuum of AI integration detected by the MLE method.

**Full-text analysis and its validations.** Our earlier analysis primarily focused on paper abstracts, but abstracts may not fully represent the writing characteristics of entire papers. Therefore, we conducted supplementary analyses on 164,579 full-text scientific publications to provide validation. These full-text documents were sourced from the complete dataset of 6,704,911 papers in OpenAlex, using the “pdf_url” field that provides links to full-text PDFs. Approximately 1.9 million papers in the dataset contained this field. Due to limitations imposed by publishers’ open access policies and database download constraints, we were unable to retrieve all available PDFs. We employed sampling and used PyMuPDF (51) for PDF parsing, ultimately obtaining full-text data. We acknowledge that the selection of full-text papers among all publications could not be fully random due to substantial PDF download failures and occasional parsing errors. However, our sampling results demonstrated stable representativeness across publication years, disciplines, and countries when compared to the complete dataset of over 6.7 million papers (Table S3).

Moreover, to further validate the representativeness of the adopted full-text publications, we calculated the occurrence of AI-characteristic keywords in each publication's full text. Results show that, between 2020 and 2022, the average number of AI-characteristic keywords per paper remained stable at approximately 15, while a notable upward trend began in 2023 (shortly after ChatGPT had released), exceeding 30 by 2025 (Fig. S2e), which is consistent with the results of the abstract analysis.

**Identification of AI usage disclosures.** To further investigate AI usage disclosure practices, we implemented a two-step analytical approach. First, we used regular expressions to identify document sections and applied a pre-defined vocabulary for pattern matching, including specific AI tools (such as "ChatGPT"), general terminology (e.g., "artificial intelligence"), and usage behavior keywords (e.g., "use" and "assist"). The regular expressions used for parsing the full text and the keywords for identifying AI disclosures are provided in Tables S5 and S6. This initial step identified text segments potentially containing AI disclosures. We observed that publications after 2023 had frequently described AI applications in experimental research rather than writing assistance; thus, our initial results included many non-writing-related AI uses. Therefore, in the second step, we input the keyword-containing paragraphs identified in the first step into LLM (see the prompt utilized in Supporting Information), which determined whether the content represented author disclosures of AI tool usage for writing purposes. This two-step process enabled us to effectively distinguish between AI applications in writing assistance versus other research contexts, thereby enhancing the reliability of our analytical findings.

**AI tool usage declaration in this work.** We declare extensive use of AI tools throughout this research. Specifically, we employed three major AI platforms: ChatGPT-4o-mini and Gemini 2.5 Flash, which were utilized primarily through API calls to handle large-scale data processing, and Claude 3.8 Sonnet, which was used via its conversational interface for qualitative auditing and logic checks. These tools were applied to three analytical tasks. First, these AI tools were used to analyze and categorize journal AI usage policies and classify them into our four-category framework. Second, for our disclosure analysis of 164k full-text papers, AI tools assisted in identifying and extracting AI usage declarations from various paper sections, including methods, acknowledgments, and author statements. Third, AI tools facilitated the task of mapping JCR's 200+ subject categories to OpenAlex's four-domain classification system. The specific prompts detailed in Supporting Information. Results were cross-validated for consistency, and all final outputs were thoroughly reviewed and validated by human researchers, maintaining the integrity and reliability of our findings.

To ensure accuracy and address potential concerns regarding technical limitations, we meticulously monitored input sizes so that all processed text remained well within the context limits of our API-based models. GPT-4o-mini and Gemini 2.5 Flash feature extensive context windows (128k tokens and up to 1 million tokens, respectively). Our granular analysis of the dataset confirms that these limits were never approached: in the manual policy collection, the three longest files each exceeded 10,000 words yet were processed as complete documents in a single API request, without approaching the context limit; in the automatically collected dataset, the longest raw text contained 9,003 words; and the 2,560 snippets derived from regular expression matching were substantially shorter, with the longest containing only 347 words. This confirms that complete contextual information was utilized for every document.

Additionally, we acknowledge that during the preparation of this paper, AI tools including Gemini-2.5, Gemini-3, and GPT-5 were used for partial language polishing. We take full responsibility for the content refined with AI assistance and have conducted a thorough verification of its accuracy.

## Acknowledgments

We thank Yihe Wang, Hongkan Chen, Jialin Liu, Zhiyu Tao, Ruijie Ma, Ruxuan Guo, and all current and former members of the Knowledge Discovery Lab at Peking University for support in empirical studies and fruitful discussions. The authors are also grateful for the constructive comments from Zhichao Fang, Hongsheng Feng, Liang Huo, Junzhi Jia, Bing Liu, Xiaomin Liu, Wen Lou, Yuntao Pan, Ming Ren, Zhixiang Wu, Bailing Zeng, Tieming Zhang, Jiuzhen Zhang, and three anonymous reviewers. This work is supported by the National Natural Science Foundation of China (#72474009) and the special project for discipline development at Peking University. We acknowledge AI use in the current work (Materials and Methods).

Supporting Information for

# Academic Journals' AI Policies Fail to Curb the Surge in AI-assisted Academic Writing

Yongyuan He[1] and Yi Bu[1,2 *]

[1] Department of Information Management, Peking University, Beijing 100871, China

[2] Center for Informationalization and Information Management Research, Peking University, Beijing 100871, China

*Corresponding author. Email: buyi@pku.edu.cn

## AI policy collection

Policy data were collected at two distinct time points. The first phase of data collection was conducted in January 2025 through a three-step manual approach:

*Step 1:* We manually searched for journal names and reviewed specific pages, including submission guidelines, author information, and journal policies, to identify texts related to policies governing authors' use of AI.

*Step 2:* We strictly excluded content concerning reviewers' use of AI or AI applications in images and focused solely on policies regarding authors' use of AI in writing.

*Step 3:* We adopted LLMs (GPT-4o-mini and Gemini 2.5 Flash) to examine and categorize the results.

The second phase took place in October 2025, employing an automated five-step crawling pipeline.

*Step 1:* We utilized the Serper API (1) to retrieve journal homepage URLs by searching for journal names combined with their ISSNs and publishers.

*Step 2:* To improve the accuracy of retrieval, we prioritized domains of well-known publishers (e.g., nature.com, sciencedirect.com, wiley.com, ieee.org, springer.com) and excluded non-journal webpages or third-party aggregators based on domain names (e.g., researchgate.net, semanticscholar.org, wikipedia.org, and repositories like sci-hub). A complete list of domain names is shown in Table S9.

*Step 3:* We employed the open-source tool Crawl4AI (2) to identify specific policy URLs, adjusting the method to task the model solely with returning target locations to mitigate token limitations and hallucinations. The identified pages were parsed, and paragraphs containing the following specific keywords as candidate AI policy content: "ai," "artificial intelligence," "llm," "generative ai," "language model," and "ChatGPT."

*Step 4:* GPT-4o-mini and Gemini 2.5 Flash were then used to identify and filter out data that did not align with our predefined policy scope. We exclusively retained texts specifying the guidelines for authors' use of AI tools during the academic writing process; policies about peer reviewers or editors, as well as those concerning AI-generated images or charts, were systematically excluded. Detailed criteria are provided in the "**Policy classification prompt**" section.

*Step 5:* For cases where websites refused automated access or were judged not to be journal webpages, we manually checked and supplemented the data.

## Publication data collection

Our publication data were primarily sourced from OpenAlex (3), a large-scale bibliographic database that covers multiple disciplines in natural sciences and engineering, life sciences, social sciences, and arts and humanities. We extracted data from the May 30, 2025 snapshot by linking JCR journals with OpenAlex journals via ISSN or journal names. We only retained English-language publication records from January 2021 to May 2025, including metadata such as DOI, title, abstract, institutional countries of all authors (Materials and Methods), and paper domains/fields. For June 2025 metadata missing from the aforementioned snapshot, we supplemented through OpenAlex's API (3), resulting in a total of 6,704,911 papers. However, we noticed that 2,768,778 records (41.29%) lacked abstracts.

To supplement missing abstracts, we retrieved corresponding records from the Web of Science database and merged them with OpenAlex records via DOI matching, using title matching (standardized to lowercase with special characters removed) as fallback. After merging and removing records with missing information, the final dataset contained 5,235,012 publications for further analyses.

## Policy classification prompt

*You are an expert in academic journal policy analysis. Given a policy text about AUTHORS' use of AI in academic writing, output a structured judgment.*

*Rules:*

*- Category (choose EXACTLY ONE of the following labels):*

*1) Strict Prohibition: clearly prohibits authors from using AI for academic writing, including editing, translation, or content generation.*

*2) Open: allows authors to use AI AND explicitly does NOT require disclosure.*

*3) Disclosure Required: allows authors to use AI BUT requires disclosure in the manuscript (extract disclosure location).*

*4) Not Mentioned: the text is not about AUTHORS' use of AI (e.g., only about peer review or editors), or insufficient to determine.*

*- Disclosure location: ONLY when category is "Disclosure Required," summarize a free-text location from the policy (do NOT choose from fixed options). Examples include but are not limited to: Methods section, Acknowledgements, Cover letter, Title page, End-of-manuscript statement, Submission system, etc. If multiple locations are acceptable, summarize briefly; if not specified, return an empty string.*

*- Reason: one or two sentences summarizing the basis for the classification; paraphrase or quote key phrases succinctly.*

*- Notes: Mentions that AI cannot be listed as an author should be treated as background and MUST NOT change the four-class decision. If the text only addresses peer review/editors' use of AI but not authors, classify as "Not Mentioned." When multiple statements exist, prioritize explicit requirements about AUTHORS' use in writing.*

*Output format (STRICT): Return ONLY a JSON object with fields:*

*{"category": one of ["Strict Prohibition", "Open", "Disclosure Required", "Not Mentioned"], "disclosure_location": <free text or empty string>, "reason": <short justification>}*

## AI disclosure recognition prompt

*You are an expert in academic publishing ethics. Please analyze the following text to identify any disclosure of AI tool usage in academic writing.*

*CRITERIA:*
*1. Focus exclusively on authors' statements about using AI tools (e.g., ChatGPT, GPT-4, Claude, Gemini, Copilot) for writing assistance, editing, or grammar correction.*
*2. Exclude cases where the paper merely researches AI tools without author usage disclosure.*
*3. Distinguish between AI tools and human names (e.g., Claude Smith).*

*4. Ignore standard methodological statements (e.g., “BERT was used for text classification”).*

*REQUIRED OUTPUT FORMAT (STRICT JSON):*
*{*
*disclosure_present: true/false,*
*disclosure_location: Methods section/Acknowledgements/Cover letter/Title page/End-of-manuscript/Submission system/Other [specify],*
*ai_tools_mentioned: [tool1, tool2, …],*
*usage_purpose: writing assistance/editing/grammar correction/content generation/translation/other,*
*rationale: brief explanation for the classification*
*}*

*ANALYSIS GUIDELINES: Infer disclosure location from context and typical academic writing conventions. Extract specific AI tool names when explicitly mentioned. Classify usage purpose based on described activities. Base all conclusions on the actual text content, not on the examples provided above. Provide concise but informative rationale.*

## Supplementary Figures

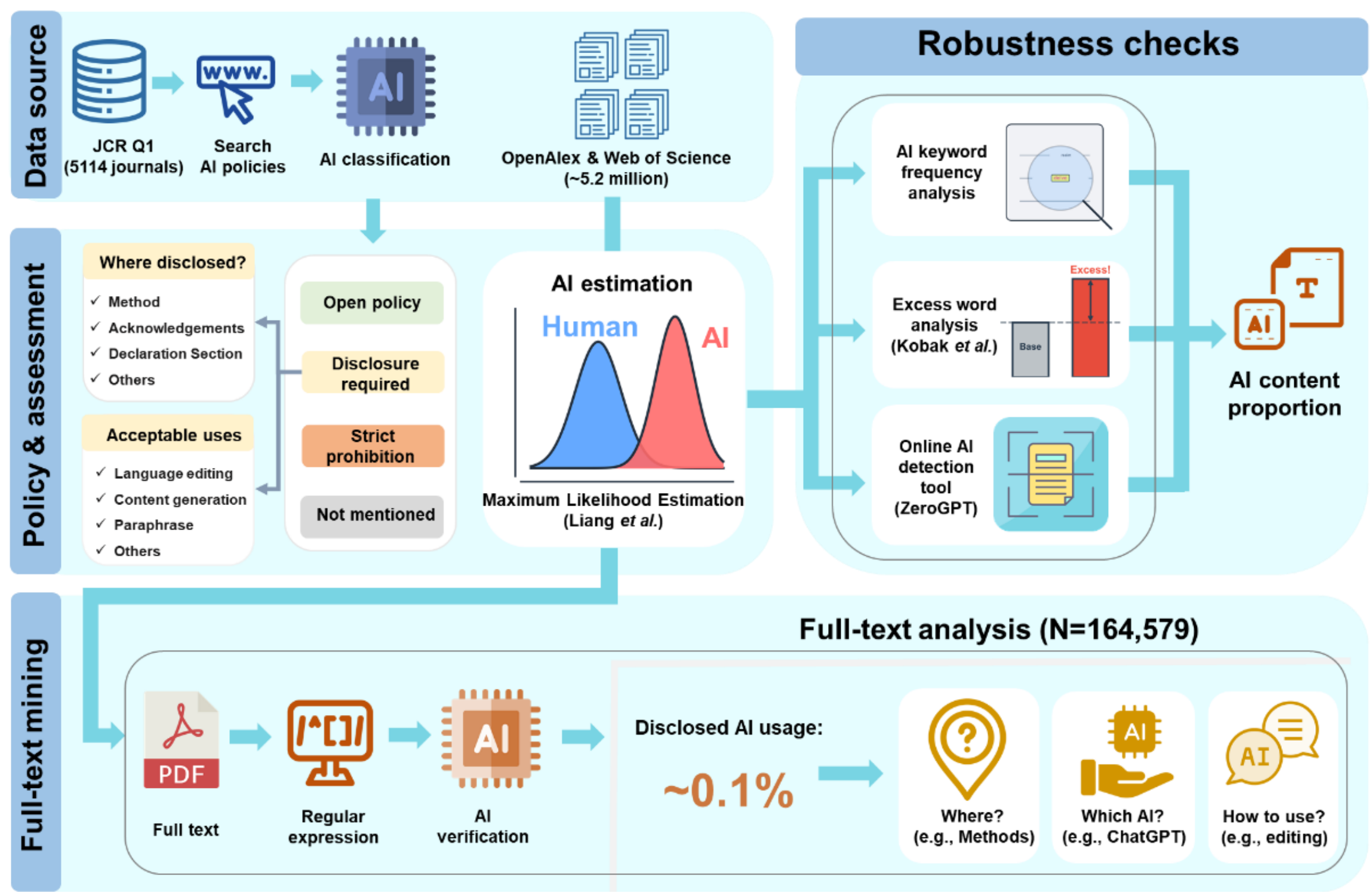

**Fig. S1. The workflow comprises three main stages**: **Stage 1: Data source and policy classification.** We analyzed 5,114 JCR Q1 journals and 5.2 million papers (abstracts) retrieved from OpenAlex and Web of Science (Materials and Methods). Journal AI policies were retrieved and classified into four categories (namely strict prohibition, open policy, disclosure required, and not mentioned) using large language models. **Stage 2: AI content estimation and robustness checks.** The primary estimation of AI content in abstracts was conducted using the Maximum Likelihood Estimation method following Liang *et al*. (4). To ensure reliability, three robustness checks were employed: AI keyword frequency analysis, excess word analysis following Kobak *et al*. (5), and independent verification using an online detection tool (ZeroGPT (6)). **Stage 3: Full-text mining.** A subset of 164,579 full-text PDFs was analyzed to estimate the rate of explicit AI disclosure. Using regular expressions and AI verification, we identified an overall disclosure rate of approximately 0.1% and extracted details regarding disclosure location (e.g., Methods), specific tools used (e.g., ChatGPT), and usage purposes (e.g., editing).

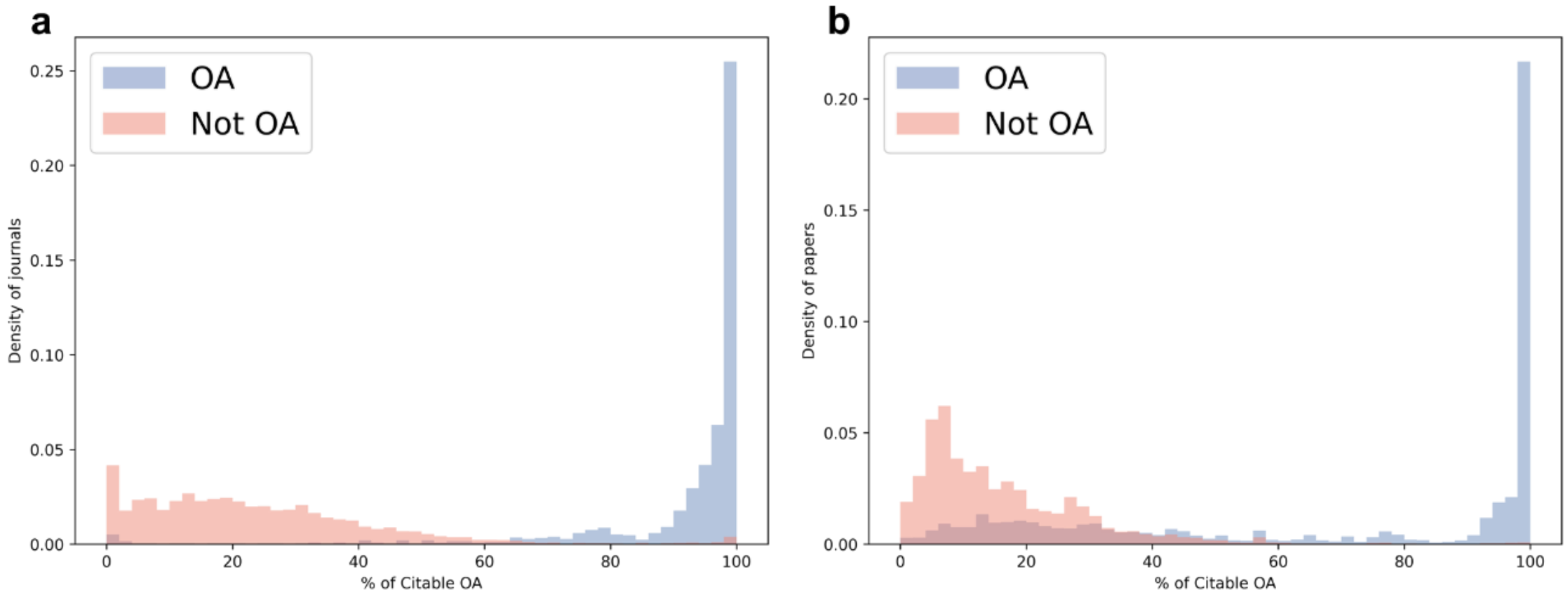

**Fig. S2. Validation of OA classification strategy based on JCR metrics. a**, Distribution at the journal level. Journals identified as Open Access by OpenAlex (blue) are predominantly clustered in the region where the "% of Citable OA" (the horizontal axis) exceeds 50%, while non-OA journals (red) are concentrated in the lower percentage region. **b**, Distribution at the publication level. The distribution pattern remains consistent when analyzing individual publications, with a clear separation between OA and non-OA publications. Descriptive statistics are shown in Tables S7 and S8.

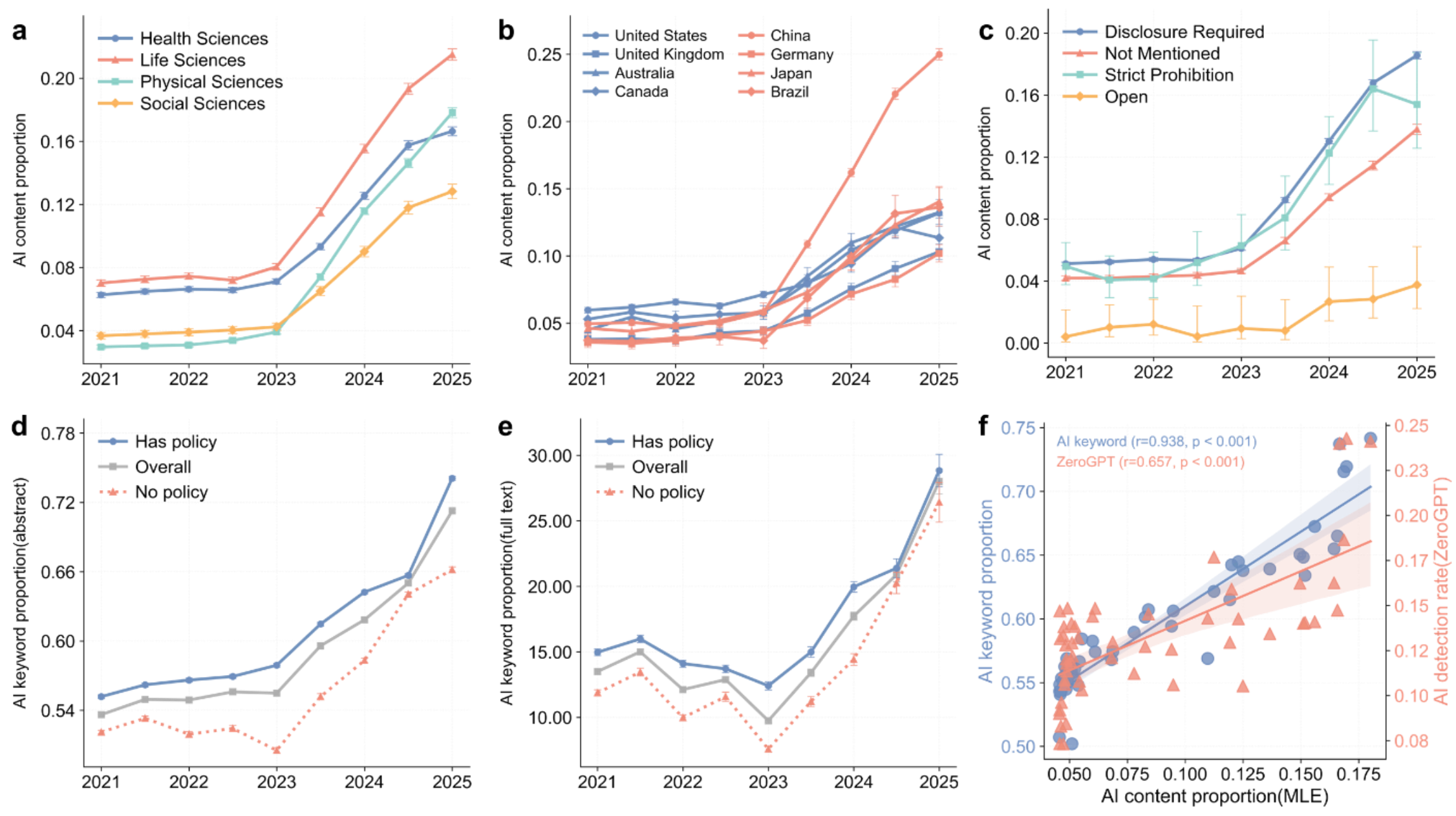

**Fig. S3. Temporal trends and correlations of AI content detection across multiple dimensions. a**, AI content proportion trends by academic domain from 2021 to 2025 across Health Sciences, Life Sciences, Physical Sciences, and Social Sciences. **b**, AI content proportion trends by language background, comparing English-speaking countries (e.g., U.S., U.K., Australia, Canada) and non-English-speaking countries (e.g., China, Germany, Japan, Brazil). Red lines represent non-English-speaking countries, blue lines represent English-speaking countries, with distinct markers for individual nations. **c**, AI content proportion trends by journal AI policy categories, including disclosure required, not mentioned, strict prohibition, and open policies. **d**, AI keyword proportion trends in abstract by journal policy status, comparing journals with AI policies, overall average, and journals without policies. **e**, AI keyword proportion trends in full text by journal policy status. **f**, Correlation between AI keyword detection and MLE-based AI content detection methods, with ZeroGPT validation results shown as a secondary comparison.

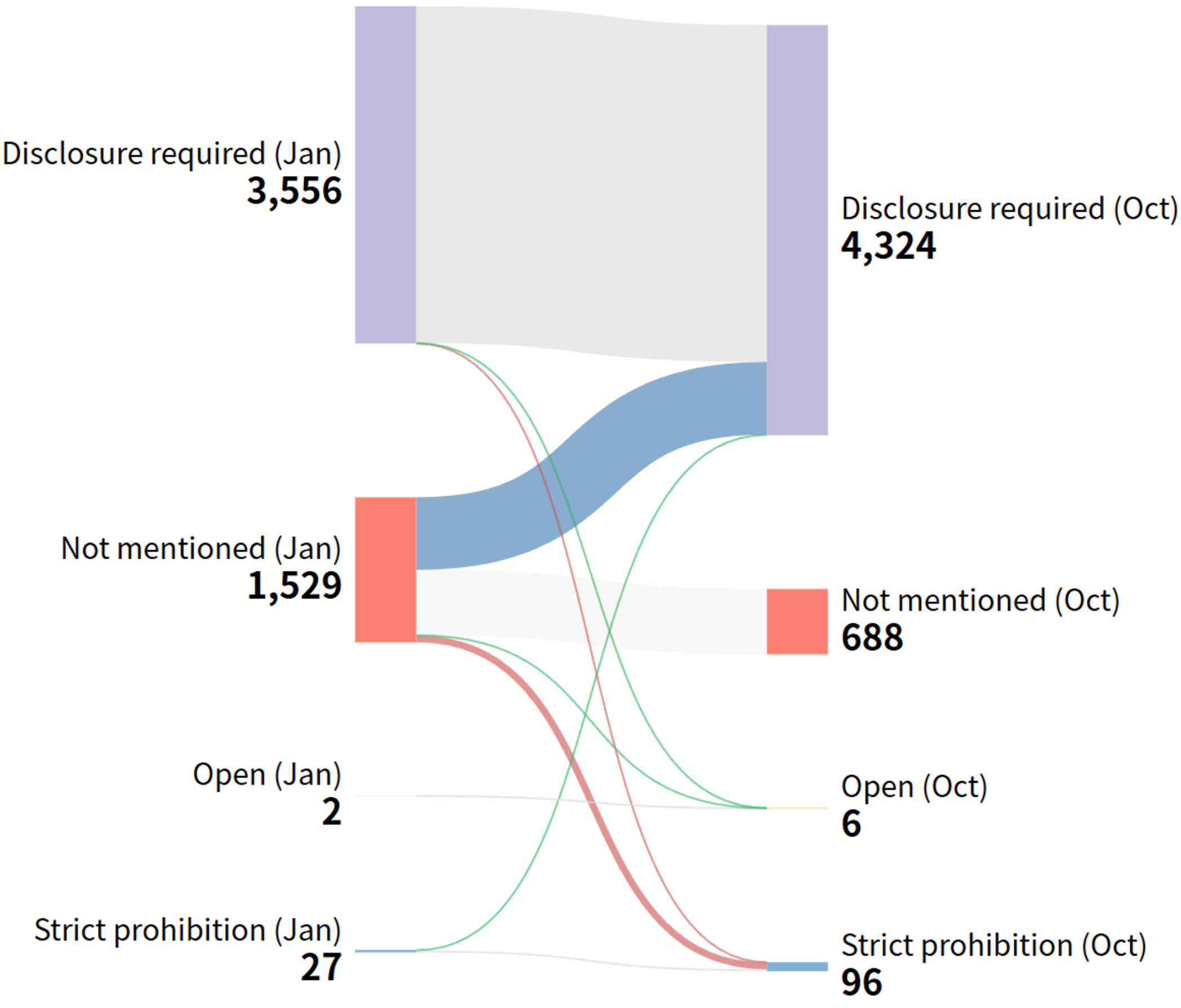

**Fig. S4. Changes in journal AI policies between January 2025** (left) **and October 2025** (right). This Sankey diagram visualizes the flow of policy changes among the 5,114 analyzed journals over a nine-month period.

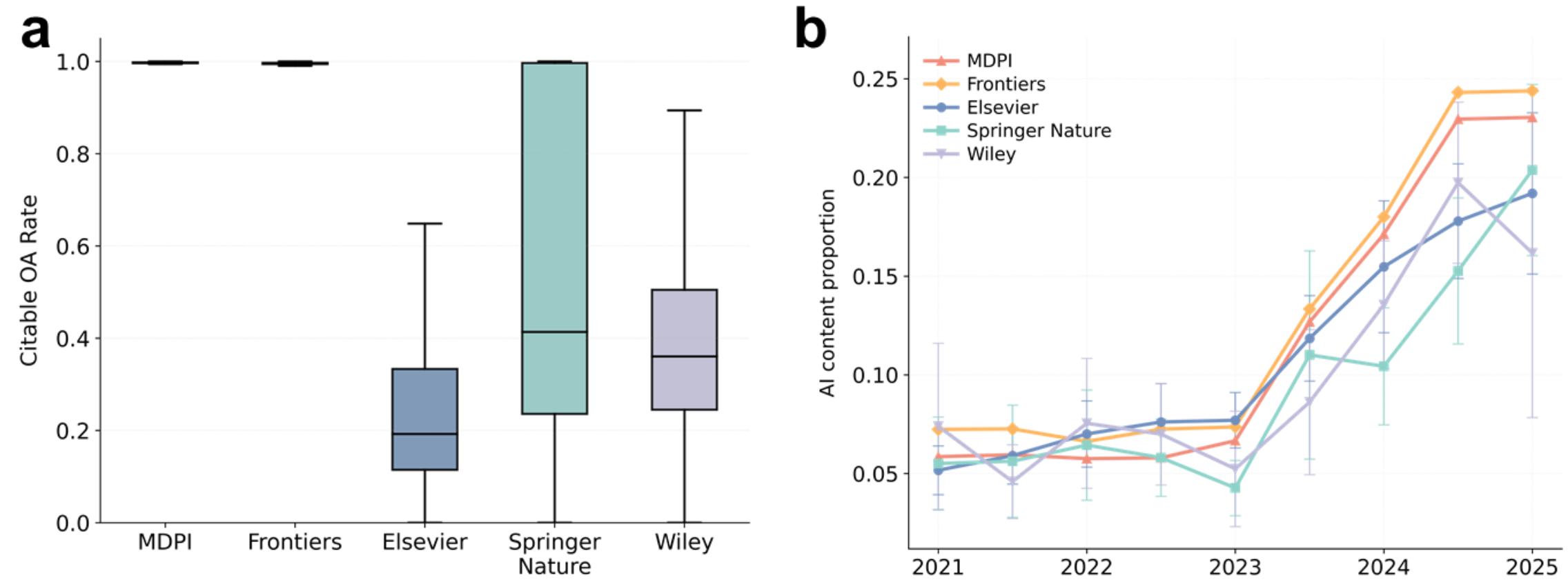

**Fig. S5. Comparison of OA and AI usage trends across major publishers. a**, Distribution of OA rates for five representative publishers (MDPI, Frontiers, Elsevier, Springer Nature, and Wiley). **b**, Temporal evolution of AI content proportion for these publishers from 2021 to 2025. Error bars represent 95% confidence intervals.

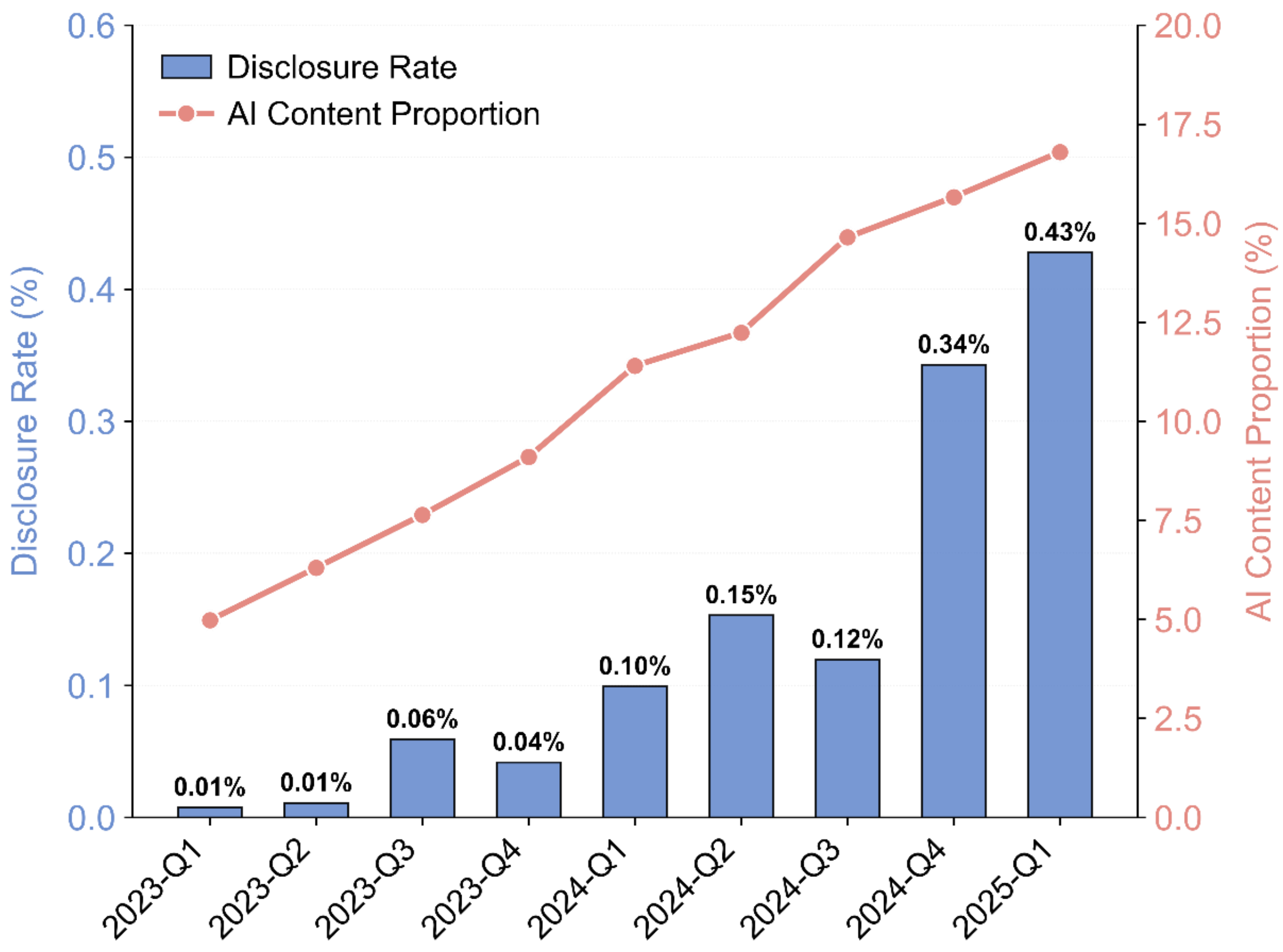

**Fig. S6. Comparison of AI disclosure rate and estimated AI usage (2023–2025).** The blue bars (left vertical axis) represent the percentage of full-text papers explicitly disclosing the use of AI tools. The red line (right vertical axis) indicates the overall AI content proportion.

## Supplementary Tables

**Table S1. Examples of journal AI policies.** The policy information presented was reviewed and recorded in January 2025.

| Policy type | Journal | Policy text |
|---|---|---|
| Open policy | Journal of Educational Evaluation for Health Professions | "The main difference from other journal publishers is that the Journal of Educational Evaluation for Health Professions (JEEHP) does not ask authors to disclose the use of AI tools. The reason for this is that the editorial office is not able to screen the use of AI tools consistently..." |
| Open policy | Journal of the Association of Environmental and Resource Economists | "Authors of papers are responsible for all content in their papers. That means the authors are solely responsible for coding errors, incorrect facts or plagiarized text arising from the use of generative AI. While it is not appropriate to cite generative AI or include such bots as co-authors, authors may include an optional disclosure statement if they want to." |
| Strict prohibition | Annals of Mathematics | "The Annals of Mathematics does not consider papers generated using AI products. Only individuals who can take full responsibility for the contents of a submission can be named authors." |
| Strict prohibition | Journal of Numerical Mathematics | "Please note that we do not accept papers that are generated by Artificial Intelligence (AI) or Machine Learning Tools primarily because such tools cannot take responsibility for the submitted work and therefore cannot be considered as authors." |
| Disclosure required | Journal of World Prehistory | "Authors who use AI tools in the writing of a manuscript, production of images or graphical elements of the paper, or in the collection and analysis of data, must be transparent in disclosing in the Materials and Methods (or similar section) of the paper how the AI tool was used and which tool was used. Authors are fully responsible for the content of their manuscript, even those parts produced by an AI tool, and are thus liable for any breach of publication ethics." |
| Disclosure required | Active Learning in Higher Education | "Disclosure: As outlined above you must clearly reveal any AI-generated content within your submission. Detail where the AI-generated content appears, using the disclosure template found at the end of these guidelines and provide this disclosure along with your submission." |

**Table S2. Sources of AI-characteristic keywords.**

<table>
<tr><th>Validation Source</th><th>AI-characteristic keywords</th></tr>
<tr><td>GPTZero (6)</td><td rowspan="6">delve, innovative, meticulous, pivotal, underscore, bolster, commendable, intricate, realm, revolutionize, shed, transformative, advancement, emphasize, endeavor, enhance, groundbreaking, harness, interplay, invaluable, multifaceted, noteworthy, nuanced, offer, renowned, seamless, strategically, streamline, unravel, valuable, comprehensive, vital, profound, adept, align, avenue, broader, burgeon, capability, compelling, comprehend, contribute, demonstrate, distinctive, elevate, elucidate, employ, encompass, endure, exceptional, explore, facilitate, foster, foundational, garner, grapple, groundwork, illuminate, imperative, inadequately, influence, inherent, integration, interconnectedness, intricacy, juxtapose, leverage, notable, nuance, pave, pioneer, poised, pose, predominantly, recognize, refine, remarkable, showcase, surpass, typically, ultimately, unlock, unparalleled, unveil, uphold, utilize, insight, significant, unwavering, crucial, highlight, importance, deeper, perspective, essential, cutting-edge, tapestry, impressively, prowess, holistic, embark, notably, seamlessly, excellently, scholarly, meticulously, tangible, methodical, fascinating, intriguing</td></tr>
<tr><td>Grammarly (7)</td></tr>
<tr><td>Delving into PubMed records: some terms in medical writing have 3 drastically changed after the arrival of ChatGPT (8)</td></tr>
<tr><td>Delving into LLM-assisted writing in biomedical publications through excess vocabulary (5)</td></tr>
<tr><td>ChatGPT “contamination”: estimating the prevalence of LLMs in the scholarly literature (9)</td></tr>
<tr><td>Delving into the Utilisation of ChatGPT in Scientific Publications in Astronomy (10)</td></tr>
</table>

**Table S3. Proportional distribution of the full-text sample across years, domains, countries, and OA status.**

| Dimension | Category | # of samples | Total count | Ratio (%) |
|---|---|---|---|---|
| Year | 2021 | 46,591 | 1,389,504 | 3.35 |
| | 2022 | 42,816 | 1,406,298 | 3.05 |
| | 2023 | 40,497 | 1,465,309 | 2.76 |
| | 2024 | 30,162 | 1,616,880 | 1.87 |
| | 2025 | 4,513 | 826,920 | 0.55 |
| Domain | Physical Sciences | 42,897 | 2,676,480 | 1.6 |
| | Health Sciences | 56,131 | 2,050,465 | 2.74 |
| | Life Sciences | 37,367 | 1,164,693 | 3.21 |
| | Social Sciences | 19,689 | 736,435 | 2.67 |
| Country | CN | 30,713 | 1,742,099 | 1.76 |
| | US | 34,823 | 1,580,867 | 2.2 |
| | GB | 16,213 | 492,688 | 3.29 |
| | DE | 13,528 | 360,008 | 3.76 |
| % of Citable OA | 0%-25% | 24,735 | 3,412,471 | 0.72 |
| | 26%-50% | 21,756 | 1,237,841 | 1.76 |
| | 51%-75% | 8,547 | 261,374 | 3.27 |
| | 76%-100% | 109,541 | 1,793,225 | 6.11 |

**Table S4. Differences in monthly growth rates (logarithmic increments) of AI content proportion between journals with and without AI policies using the Mann-Whitney U test.**

| Subgroup | mean growth(has_policy) | mean growth(no_policy) | U | *p*-value |
|---|---|---|---|---|
| **Overall** | 0.042 | 0.059 | 411 | 0.722 |
| **By language** | | | | |
| English | 0.077 | -0.020 | 470 | 0.446 |
| Non-English | 0.089 | 0.152 | 408 | 0.852 |
| **By domain** | | | | |
| Social Science | 0.038 | 0.049 | 417 | 0.963 |
| Life Science | 0.034 | 0.037 | 405 | 0.816 |
| Physical Sciences | 0.054 | 0.056 | 405 | 0.816 |
| Health Sciences | 0.031 | 0.024 | 456 | 0.586 |
| **By OA status** | | | | |
| High OA | 0.041 | 0.046 | 403 | 0.792 |
| Low OA | 0.042 | 0.037 | 410 | 0.876 |

**Table S5. Regular expressions used for PDF parsing.**

<table>
<tr><th>Category</th><th>Target/Section</th><th>Regular expression pattern</th></tr>
<tr><td rowspan="8"><b>Text cleaning</b></td><td>References removal</td><td>(references|bibliography)\s*\n.*?($|\Z)</td></tr>
<tr><td rowspan="3">Metadata (Header/footer)</td><td>\n\s*\d+\s*\n (Page numbers)</td></tr>
<tr><td>\n.{0,50}Vol\.\s*\d+.{0,30}\n (Volume info)</td></tr>
<tr><td>\n.{0,50}20\d{2}.{0,30}\n (Year info)</td></tr>
<tr><td rowspan="2">URLs & DOIs</td><td>doi:?\s*10\.\d+/\S+</td></tr>
<tr><td>https?://\S+</td></tr>
<tr><td rowspan="2">Format normalization</td><td>{2,} (Multiple spaces)</td></tr>
<tr><td>\n{3,} (Excessive newlines)</td></tr>
<tr><td rowspan="15"><b>Structure segmentation</b></td><td>Introduction</td><td>(introduction|preface)[\s\.:]*</td></tr>
<tr><td rowspan="3">Methods / experimental</td><td>method(s|ology)?[\s\.:]*</td></tr>
<tr><td>experimental[\s\w]*section</td></tr>
<tr><td>materials[\s\w]*(and|&)[\s\w]*methods</td></tr>
<tr><td rowspan="3">Ethics & disclosure</td><td>ethic(s|al)?[\s\w]*statement</td></tr>
<tr><td>disclosure(s)?[\s\.:]*</td></tr>
<tr><td>declaration[\s\w]*of[\s\w]*interest(s)?</td></tr>
<tr><td rowspan="3">Acknowledgements/funding</td><td>acknowledge[a-z]*[\s\.:]*</td></tr>
<tr><td>funding[\s\.:]*</td></tr>
<tr><td>grant(s)?[\s\.:]*</td></tr>
<tr><td rowspan="2">AI declaration</td><td>(ai|llm)[\s\w]*</td></tr>
<tr><td>(declaration|statement|disclosure)</td></tr>
<tr><td rowspan="2">Figure captions</td><td>figure[\s\w]*caption(s)?</td></tr>
<tr><td>figure[\s\w]*legend(s)?</td></tr>
</table>

**Table S6. Keywords and regular expressions related to AI disclosure.**

<table>
<tr><th>Category</th><th>Sub-category</th><th>Keywords or regex patterns</th></tr>
<tr><td rowspan="3"><b>AI terminology</b></td><td>LLM families</td><td>\b(gpt(-[34]\.?[5]?)?|gpt[34]|chatgpt|claude(-\d)?|anthropic|bard|gemini|palm(-2)?|llama(-\d)?|meta\s+ai|mistral|mixtral|falcon|openai|microsoft\s+copilot|bing\s+(chat|ai))\b</td></tr>
<tr><td>General terms</td><td>\b(large\s+language\s+model|llm|language\s+model|generative\s+ai|foundation\s+model|ai(\s+writing)?\s+assistant|ai\s+tool|generative\s+model)\b</td></tr>
<tr><td>Writing tools</td><td>\b(grammarly|quillbot|writesonic|jasper|copy\.ai|wordtune|hemingway|paperpal|scite|trinka|writefull|scholarcy|scispace|elicit)\b</td></tr>
<tr><td rowspan="12"><b>Disclosure patterns</b></td><td rowspan="2">Active declaration</td><td>(used|utilized|employed|leveraged|applied)[\s\w]{0,30}(ai|artificial\s+intelligence|language\s+model|llm|generative|assistant)</td></tr>
<tr><td>(assisted|supported|aided|helped|generated|created|drafted|written|edited|revised|proofread)[\s\w]{0,30}(by|with|using)[\s\w]{0,30}(ai|artificial\s+intelligence|language\s+model|llm|generative|assistant)</td></tr>
<tr><td rowspan="2">Passive declaration</td><td>(ai|artificial\s+intelligence|language\s+model|llm|generative|assistant)[\s\w]{0,30}(was|were|has\s+been|have\s+been)[\s\w]{0,30}(used|utilized|employed|leveraged|applied)</td></tr>
<tr><td>(manuscript|text|writing|draft|paper|article|content|language|grammar)[\s\w]{0,30}(was|were|has\s+been|have\s+been)[\s\w]{0,30}(assisted|supported|generated|checked|improved|enhanced|refined)[\s\w]{0,30}(by|with|using)[\s\w]{0,30}(ai|artificial\s+intelligence|language\s+model|llm|generative|assistant)</td></tr>
<tr><td rowspan="2">Acknowledgment</td><td>(acknowledg|thank|gratitude|grateful)[\s\w]{0,50}(ai|artificial\s+intelligence|language\s+model|llm|generative|assistant)</td></tr>
<tr><td>(acknowledg|thank|gratitude|grateful)[\s\w]{0,50}(support|assistance|help|contribution)[\s\w]{0,30}(from|by|of)[\s\w]{0,30}(ai|artificial\s+intelligence|language\s+model|llm|generative|assistant)</td></tr>
<tr><td rowspan="2">Explicit statement</td><td>this[\s\w]{0,20}(paper|manuscript|article|research|study|work)[\s\w]{0,30}(uses|used|utilizes|utilized)[\s\w]{0,30}(ai|artificial\s+intelligence|language\s+model|llm|generative|assistant)</td></tr>
<tr><td>this[\s\w]{0,20}(paper|manuscript|article|research|study|work)[\s\w]{0,30}(is|was|has\s+been)[\s\w]{0,30}(ai-assisted|ai-generated|ai-written|ai-enhanced)</td></tr>
<tr><td rowspan="2">Negative statement</td><td>(no|not)[\s\w]{0,20}(ai|artificial\s+intelligence|language\s+model|llm|generative|assistant)[\s\w]{0,30}(was|were)[\s\w]{0,30}(used|utilized|employed)</td></tr>
<tr><td>(did\s+not|didn't|have\s+not|haven't)[\s\w]{0,20}(use|employ|utilize)[\s\w]{0,30}(ai|artificial\s+intelligence|language\s+model|llm|generative|assistant)</td></tr>
<tr><td rowspan="2">Other statements</td><td>(only|just|specifically)[\s\w]{0,30}(used|utilized|employed)[\s\w]{0,30}(ai|artificial\s+intelligence|language\s+model|llm|generative|assistant)[\s\w]{0,30}(for|to)</td></tr>
<tr><td>(ai|artificial\s+intelligence|language\s+model|llm|generative|assistant)[\s\w]{0,30}(was|were)[\s\w]{0,30}(only|just|specifically)[\s\w]{0,30}(used|utilized|employed)[\s\w]{0,30}(for|to)</td></tr>
</table>

### Table S7. Cross-validation of journal open access status between JCR and OpenAlex

| | **OpenAlex classification** | |
|---|---|---|
| **JCR: % of Citable OA** | **Non-OA journals** | **OA journals** |
| **0%-25%** | 2,310 (45.08%) | 25 (0.49%) |
| **26%-50%** | 1,252 (24.43%) | 23 (0.45%) |
| **51%-75%** | 216 (4.22%) | 70 (1.37%) |
| **76%-100%** | 77 (1.50%) | 1,151 (22.46%) |

### Table S8. Cross-validation of publication open access status between JCR and OpenAlex

| | **OpenAlex classification** | |
|---|---|---|
| **JCR: % of Citable OA** | **Non-OA publications** | **OA publications** |
| **0%-25%** | 1,942,393 (37.10%) | 562,707 (10.75%) |
| **26%-50%** | 472,238 (9.02%) | 407,891 (7.79%) |
| **51%-75%** | 48,531 (0.93%) | 171,779 (3.28%) |
| **76%-100%** | 23,083 (0.44%) | 1,606,390 (30.69%) |

### Table S9. List of domains used for filtering during the process of obtaining journal homepage URLs

| Category | List of domains (ordered alphabetically) |
|---|---|
| **Publishers** | academic.oup.com, acm.org, aip.org, amegroups.com, bjournals.org, cambridge.org, cell.com, elsevier.com, frontiersin.org, hindawi.com, ieee.org, iop.org, journals.aps.org, kluweronline.com, mdpi.com, nature.com, oup.com, oxfordjournals.org, rsc.org, sagepub.com, sciencedirect.com, springer.com, tandfonline.com, wiley.com |
| **Non-publisher webpages** | baike.baidu.com, cnki.net, core.ac.uk, crossref.org, dblp.org, issn.org, readcube.com, researchgate.net, scholar.google.com, sci-hub, scimagojr.com, semanticscholar.org, wanfangdata.com, wikipedia.org, worldcat.org |

## Supplementary References